\documentclass[runningheads]{llncs}
\usepackage[T1]{fontenc}
\usepackage{graphicx,verbatim}
\usepackage{amsmath}
\usepackage{amssymb}
\usepackage{booktabs}
\usepackage{pifont}
\usepackage{multirow}
\usepackage{array}
\begin{document}
\title{Decouple and Reason: Anatomically Guided Two-Stage Voxel-Level Grounding of Free-Text Findings in 3D Chest CT}
%
\author{First Author\inst{1}\orcidID{0000-1111-2222-3333} \and
Second Author\inst{2,3}\orcidID{1111-2222-3333-4444} \and
Third Author\inst{3}\orcidID{2222--3333-4444-5555}}
\authorrunning{F. Author et al.}
%

\author{Kwang-Hyun Uhm\inst{1}\thanks{Corresponding author.} \and 
Inhwa Son\inst{1} \and    
Sung-Jea Ko\inst{2}}   
\authorrunning{K.-H. Uhm et al.}
%
\institute{Department of Artificial Intelligence, Gachon University, Republic of Korea 
\email{\{khuhm, inhwa1127\}@gachon.ac.kr} \and
MEDAI, Republic of Korea \\
\email{sjko@medai.im}}


  
\maketitle              
\begin{abstract}
Automatic voxel-level grounding of free-text findings in 3D chest Computed Tomography (CT) is critical for clinical interpretability. 
However, this task remains highly challenging due to the intricate spatial complexity of large 3D volumes and the heterogeneity of free-text findings. 
Existing end-to-end approaches often struggle to simultaneously learn the localized feature representations required for accurate 3D segmentation and the complex semantic understanding needed for text alignment, leading to suboptimal grounding performance. To overcome this fundamental limitation, we propose a novel decoupled framework that disentangles the problem into two specialized stages: (1) class-agnostic lesion segmentation and (2) text-volume reasoning. 
This structural separation allows the model to first extract candidate sub-volumes by localizing potential abnormalities.
Subsequently, intensive cross-modal reasoning is performed to align these localized sub-volumes with free-text medical findings. 
To resolve the spatial ambiguities inherent in local regions, the reasoning module is augmented with explicit anatomical guidance, utilizing relative spatial coordinates and lung lobe priors. Evaluated on the ReXGroundingCT benchmark, our method achieves state-of-the-art performance in overall grounding quality on the official leaderboard.
These results demonstrate that decoupling detection from reasoning is a highly effective paradigm for handling the complexity of 3D medical visual grounding. Our code is publicly available at https://github.com/khuhm/DAGG.


\end{abstract}

\section{Introduction}
Radiology reports serve as the primary medium for communicating clinical findings from 3D Computed Tomography (CT) scans~\cite{hamamci2024ctrate}. However, these free-text reports inherently lack direct spatial mapping to the visual anatomy, limiting their utility for intuitive clinical interpretation and downstream automated analysis~\cite{baharoon2025rexgroundingct}. To bridge this gap, 3D visual grounding, the task of localizing and segmenting specific regions in a 3D volume based on natural language queries, has emerged as a crucial area of research~\cite{Zhao2025sat}. In the medical domain, advancing this from coarse bounding boxes to precise voxel-level segmentation masks is of paramount importance for developing reliable Computer-Aided Diagnosis (CAD) systems~\cite{rokuss2025voxtell}.

Driven by recent advances in Vision-Language (VL) models, predominantly in the 2D domain, emerging research has begun to explore text-prompted volumetric segmentation in 3D medical images~\cite{du2024segvol,zhao2024biomedparse,Zhao2025sat,rokuss2025voxtell}. However, current methodologies often rely on simplistic, anatomical keywords (e.g., ``lung'' or ``kidney''). Consequently, they exhibit fundamental limitations in comprehending and grounding highly descriptive, free-text abnormal findings, such as ``subpleural consolidation area with air bronchograms in the left lower lobe superior segment.''

Furthermore, adapting existing end-to-end frameworks to tackle these complex queries presents significant challenges.
Medical free-text findings exhibit considerable complexity and heterogeneity~\cite{jain2021radgraph}.
Attempting to simultaneously optimize a single network for both precise 3D spatial localization and complex free-text reasoning imposes a substantial representational burden.
This makes it exceptionally difficult to learn the precise cross-modal mappings required for accurate grounding. Consequently, state-of-the-art end-to-end approaches suffer from suboptimal performance, frequently failing to either detect subtle lesions or correctly align them with the corresponding text~\cite{baharoon2025rexgroundingct}.

To overcome these fundamental limitations, we propose a novel paradigm: a Decoupled Anatomically-Guided Grounding (DAGG) Framework that disentangles the complex grounding task into two specialized, manageable stages. In the first stage, we perform a class-agnostic lesion segmentation. By abstracting away the pathological and linguistic heterogeneity, the network acts as a universal lesion detector focused solely on localizing potential abnormalities. 
This stage extracts instance-level candidate sub-volumes, allowing the subsequent module to focus entirely on cross-modal reasoning.
In the second stage, localized text-volume reasoning is intensively performed on the extracted sub-volumes. 
To enrich these local representations with global anatomical context, we introduce an Explicit Anatomical Guidance mechanism, augmenting each sub-volume with relative spatial coordinates and lung lobe priors. 
Furthermore, we identify that standard contrastive learning formulations (e.g., InfoNCE~\cite{oord2018representation}), which dominate general VL tasks, are intrinsically unsuited for medical grounding. Medical reports frequently contain semantic overlaps (e.g., identical ``small nodule in the left upper lobe'' findings across different patients), which contrastive losses incorrectly penalize as negative pairs. Furthermore, false-positive candidates generated in the first stage lack ground-truth text, making positive pairing impossible. To address this, we introduce an Independent Pairwise Alignment strategy. By optimizing the alignment score of each image-text pair independently, our model effectively suppresses false positives without distorting the latent space through false-negative sampling.
Our main contributions are summarized as follows:
\begin{itemize}
    \item We propose a novel decoupled two-stage framework for voxel-level 3D chest CT grounding, which explicitly disentangles lesion localization from semantic alignment to achieve precise cross-modal reasoning.
    \item We introduce an explicit anatomical guidance mechanism to resolve spatial ambiguities in local sub-volumes, coupled with an independent pairwise alignment strategy designed to robustly optimize cross-modal mappings given the semantic complexities of free-text medical findings.
    \item Evaluated on the recently released ReXGroundingCT~\cite{baharoon2025rexgroundingct} benchmark, our method achieves state-of-the-art performance in overall grounding quality, setting a new standard on the official leaderboard.
\end{itemize}

\begin{figure}[t]
\includegraphics[]{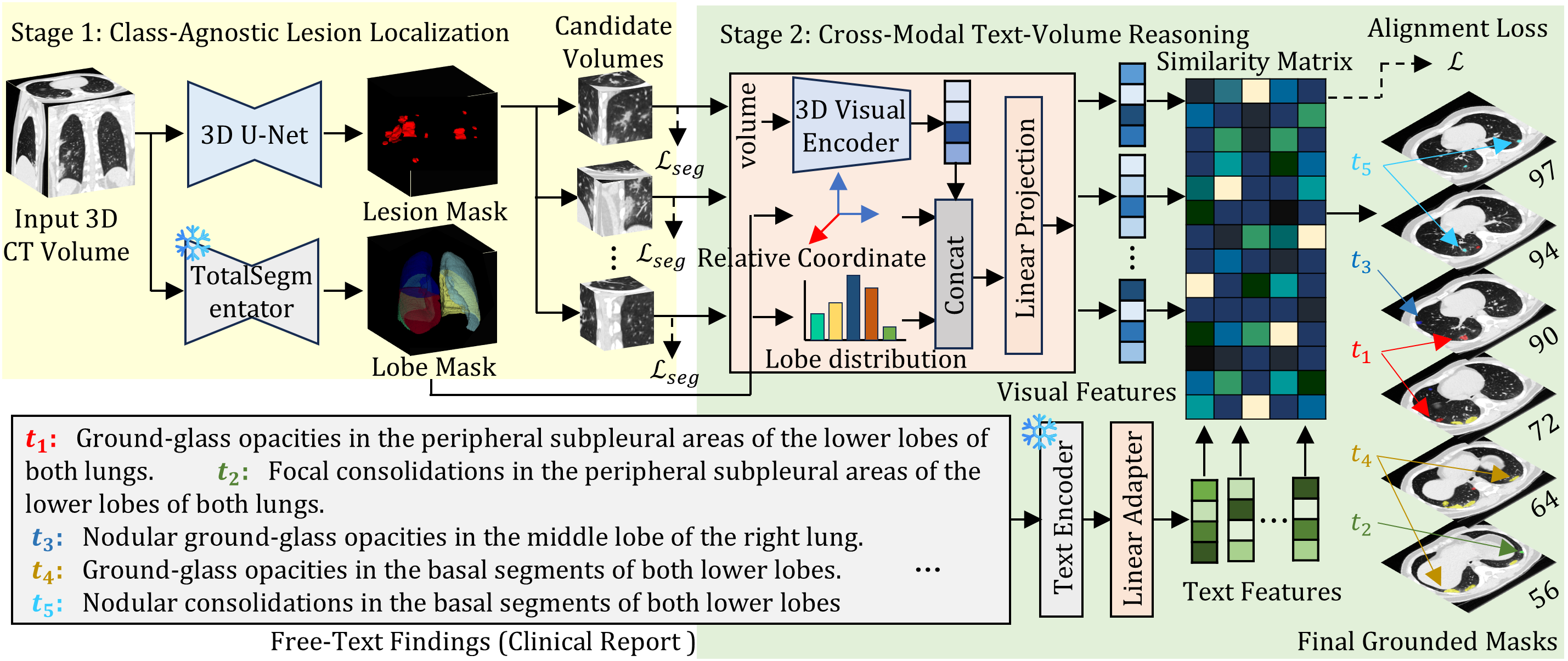}
\caption{Overall architecture of the proposed DAGG framework. Stage 1 localizes class-agnostic candidate lesions and parses anatomy, while Stage 2 aligns text embeddings with anatomically-enriched visual features via independent pairwise alignment.} \label{fig1}
\end{figure}

\section{Method}
\subsection{Problem Formulation and Overall Architecture}
\subsubsection{Problem Formulation.}
Let $V \in \mathbb{R}^{H \times W \times D}$ denote an input 3D CT volume, where $H$, $W$, and $D$ represent the spatial dimensions (height, width, and depth). Given a corresponding radiology report comprising $N$ descriptive free-text findings, denoted as $T = \{t_1, t_2, \dots, t_N\}$, the objective of our 3D medical visual grounding task is to predict voxel-level segmentation masks $M \in \{0, 1\}^{N \times H \times W \times D}$. Specifically, for each free-text finding $t_i$, the model aims to generate a precise binary mask $M_i \in \{0, 1\}^{H \times W \times D}$ that localizes the described abnormality within the 3D volume.
\subsubsection{Overall Architecture.} 
As illustrated in Figure~\ref{fig1}, we propose a novel decoupled framework that alleviates the representational burden of standard end-to-end approaches by disentangling the grounding process into two specialized stages.
In the first stage, the model comprehensively scans the entire volume $V$ to localize and extract candidate sub-volumes containing potential abnormalities. Subsequently, in the second stage, intensive cross-modal reasoning is performed between these localized sub-volumes and the free-text queries $T$ to align the visual features with the semantic context, ultimately yielding the final grounded masks $M$.

\subsection{Stage 1: Class-Agnostic Lesion Localization}
The primary objective of the first stage is to comprehensively detect all potential abnormalities within the input volume $V$, relying exclusively on visual features without any text intervention. To achieve this, we formulate the initial detection as a class-agnostic segmentation task. Specifically, rather than predicting separate masks for each finding, we merge all available ground-truth masks $M \in \{0, 1\}^{N \times H \times W \times D}$ across the $N$ findings into a single, unified binary lesion mask $M_{lesion} \in \{0, 1\}^{H \times W \times D}$. This unified formulation encourages the network to learn generalized visual features across diverse lesion types. We adopt a standard 3D U-Net~\cite{cicek20163d} for this stage, trained following the nnU-Net framework~\cite{isensee2021nnu}.

After training,  the network generates a predicted binary lesion mask $\hat{M}_{lesion}$ for a given input volume $V$. To extract individual abnormalities for precise instance-level reasoning, we apply 3D connected component analysis (CCA) to $\hat{M}_{lesion}$, identifying distinct lesion instances. Subsequently, we compute the spatial centroid of each instance. Using these centroids as anchor points, we crop local 3D sub-volumes of a fixed patch size directly from the original volume $V$. This process yields a total of $K$ candidate lesion volumes, formalized as a 4D tensor $V_{cand} \in \mathbb{R}^{K \times H_p \times W_p \times D_p}$, where $H_p$, $W_p$, and $D_p$ denote the spatial dimensions of the patches. These extracted volumes are thereby prepared as the primary visual inputs for the cross-modal reasoning in Stage 2.

Furthermore, to explicitly facilitate spatial reasoning in the downstream phase, we extract anatomical lung lobe masks from the input volume $V$. Specifically, we generate a multi-class lobe mask $M_{lobe} \in \{0, 1\}^{5 \times H \times W \times D}$ encompassing five distinct regions: the left upper, left lower, right upper, right middle, and right lower lobes. For this anatomical parsing, we employ the off-the-shelf TotalSegmentator tool \cite{wasserthal2023totalsegmentator}, providing robust spatial priors to guide the cross-modal alignment.

\subsection{Stage 2: Cross-Modal Text-Volume Reasoning}
\subsubsection{Feature Encoding and Explicit Anatomical Guidance}
In this stage, we align the extracted candidate volumes $V_{cand}$ with the free-text queries $T$ within a shared latent space.
We first encode the free-text queries $T$ using a frozen pre-trained text encoder. Each textual finding $t_i \in T$ is encoded into a feature vector $h_i \in \mathbb{R}^{d_{txt}}$, where $d_{txt}$ is the hidden dimension of the encoder. A trainable linear adapter then projects these embeddings into a shared multi-modal latent space, yielding the final text features $Z_{txt} \in \mathbb{R}^{N \times d}$, where $d$ is the joint embedding dimension.

For the visual modality, we process the candidate volumes $V_{cand}$ through a 3D CNN-based image encoder to extract local image embeddings. However, these local features inherently lack the global positional context necessary to align with spatially descriptive text findings (e.g., ``consolidation in the left lower lobe''). To bridge this spatial gap, we introduce explicit anatomical guidance. 
First, we compute a relative coordinate vector $v_{coord} \in [-1, 1]^3$ representing the 3D centroid of each candidate volume, normalized with respect to the global bounding box of the lung foreground. 
Second, we extract a lobe distribution vector $v_{lobe} \in [0, 1]^5$, quantifying the volumetric fraction of each of the five lung lobes occupying the candidate sub-volume.
These two anatomical vectors are concatenated with the local image embedding to form an anatomically-augmented visual feature. Finally, a linear projection maps this concatenated representation into the shared latent space, yielding the final visual features $Z_{vis} \in \mathbb{R}^{K \times d}$.

\subsubsection{Independent Pairwise Alignment}
Finally, we compute the pairwise cosine similarity between $Z_{vis}$ and $Z_{txt}$, yielding a dense similarity matrix $S \in \mathbb{R}^{K \times N}$. Each element $S_{i,j}$ represents the similarity score between the $i$-th visual feature $v_i \in Z_{vis}$ and the $j$-th text feature $z_j \in Z_{txt}$, calculated as  $S_{i,j} = \frac{v_i \cdot z_j}{\|v_i\| \|z_j\|}$. 
During training, we formulate the matching as an independent pairwise alignment task. For each pair $(v_i, z_j)$, we assign a ground-truth label $y_{i,j} \in \{1, -1\}$. Specifically, $y_{i,j} = 1$ if the lesion mask of $v_i$ spatially overlaps with the ground-truth mask corresponding to $t_j$; otherwise, $y_{i,j} = -1$. Notably, this independent formulation naturally accommodates candidate volumes that do not correspond to any text query, reflecting the clinical reality of false-positive localizations from the first stage.
To establish precise, independent regional correspondences for each textual finding, we optimize the network using a Cosine Embedding Loss $\mathcal{L}$ applied to each pair $(v_i, z_j)$, defined as 

$$\mathcal{L}_{i,j} = \begin{cases} 1 - S_{i,j}, & \text{if } y_{i,j} = 1 \\ \max(0, S_{i,j} - m), & \text{if } y_{i,j} = -1 \end{cases},$$
where $m \in [-1, 1]$ is a predefined margin. The overall objective is computed as the average of the mean positive loss and the mean negative loss across all $K \times N$ pairs. During inference, a pair $(v_i, z_j)$ is considered a valid match if $S_{i,j} > \tau$, where $\tau$ is a fixed threshold.

\subsubsection{Auxiliary Local Mask Refinement}
Additionally, to explicitly refine the local mask boundaries within each candidate sub-volume, we employ a separate, lightweight 3D U-Net. This localized module is supervised by an auxiliary segmentation loss $\mathcal{L}_{seg}$, trained independently from the main cross-modal matching objective.

\section{Experiments}
\subsection{Experimental Setup}
\subsubsection{Datasets} We evaluate our proposed method on ReXGroundingCT~\cite{baharoon2025rexgroundingct}, the first publicly available dataset that establishes explicit linkages between free-text radiological findings and pixel-level 3D segmentations. The dataset comprises 3,142 non-contrast chest CT studies paired with standardized radiology reports derived from CT-RATE~\cite{hamamci2024ctrate}. 
CT volumes exhibit varying physical dimensions (spacing: 0.30--0.98 mm, thickness: 0.50--5.00 mm, slices: 104--1005) with 92.6\% reconstructed at 512$\times$512 resolution.
ReXGroundingCT provides 16,301 annotated entities across 8,028 text-to-3D-segmentation pairs. Approximately 79\% of the findings are focal abnormalities, while the remaining 21\% are non-focal. The dataset includes a public validation set of 50 cases and a private test set of 100 cases, both annotated by board-certified radiologists. Private test set results are available on the official leaderboard.\footnote{\url{https://rexrank.ai/ReXGroundingCT}}.

\subsubsection{Implementation Details}
In Stage 1, the candidate localization model is trained using nnU-Net~\cite{isensee2021nnu}. CT volumes are resampled to $0.7 \times 0.7 \times 1.0$ mm$^3$, intensity-clipped to $[-999, 205]$ HU, and normalized. In Stage 2, candidate sub-volumes are extracted as $80 \times 80 \times 48$ patches. We use Qwen3-Embedding-4B~\cite{zhang2025qwen3} (prepended with the instruction ``describe the given CT findings'') and 3D SegResNet~\cite{myronenko20183d} as text and visual encoders, respectively. Their feature dimensions are 2560 and 1024, both mapped to a joint embedding space $d=2560$. A 4-level lightweight 3D U-Net handles auxiliary mask refinement. Stage 2 is optimized for 100 epochs via Adam ($lr=1 \times 10^{-3}$, batch size 32) with standard augmentations. We set $m=0$ and $\tau=0.4$, conducting all experiments on a single NVIDIA H100 GPU.

\subsubsection{Evaluation Metrics}
We evaluate the overall segmentation quality using the Global Dice score. We also report the Global HIT Rate (HIT$_{10\%}$), the proportion of findings with Dice $\ge$ 0.1. Additionally, we report instance-level Precision, Recall, and F1 score, where a true positive requires a Dice $\ge$ 0.2.

\subsubsection{Baselines}
We compare our approach against four recent state-of-the-art text-prompted models: BiomedParseV2 \cite{zhao2024biomedparse}, SegVol \cite{du2024segvol}, SAT \cite{Zhao2025sat}, and VoxTell \cite{rokuss2025voxtell}. For fair comparison, all baselines were fine-tuned on the training split.

\begin{table*}[t]
\centering
\caption{Performance on the ReXGroundingCT public validation set.}
\label{tab:val_results}
\begin{tabular}{l *{5}{>{\centering\arraybackslash}p{1.6cm}}}
\toprule
Method & Dice $\uparrow$ & F1 $\uparrow$ & Precision $\uparrow$ & Recall $\uparrow$ & HIT$_{10\%}$ $\uparrow$ \\
\midrule
BiomedParseV2~\cite{zhao2024biomedparse} & 0.059 & 0.072 & 0.061 & 0.089 & 0.147 \\
SegVol~\cite{du2024segvol}             & 0.065 & 0.081 & 0.070 & 0.097 & 0.198 \\
SAT~\cite{Zhao2025sat}                 & 0.139 & 0.166 & 0.123 & 0.259 & 0.400 \\
VoxTell~\cite{rokuss2025voxtell}       & 0.214 & 0.192 & 0.160 & 0.240 & 0.487 \\
\midrule
\textbf{DAGG (Ours)}                   & \textbf{0.250} & \textbf{0.237} & \textbf{0.196} & \textbf{0.298} & \textbf{0.565} \\
\bottomrule
\end{tabular}
\end{table*}

\begin{table*}[t]
\centering
\caption{Performance on the ReXGroundingCT private test set (Official Leaderboard).}
\label{tab:test_results}
\begin{tabular}{l *{5}{>{\centering\arraybackslash}p{1.6cm}}}
\toprule
Method & Dice $\uparrow$ & F1 $\uparrow$ & Precision $\uparrow$ & Recall $\uparrow$ & HIT$_{10\%}$ $\uparrow$ \\
\midrule
SAT~\cite{Zhao2025sat}                   & 0.000 & 0.000 & 0.000 & 0.000 & 0.000 \\
SegVol~\cite{du2024segvol}                & 0.000 & 0.000 & 0.000 & 0.000 & 0.000 \\
BiomedParseV2~\cite{zhao2024biomedparse} & 0.025 & 0.012 & 0.007 & 0.072 & 0.066 \\
SAT-FT~\cite{Zhao2025sat}                & 0.209 & 0.123 & 0.074 & \textbf{0.369} & 0.473 \\
\midrule
\textbf{DAGG (Ours)}                      & \textbf{0.253} & \textbf{0.247} & \textbf{0.211} & 0.299 & \textbf{0.517} \\
\bottomrule
\end{tabular}
\end{table*}

\subsection{Quantitative Results}

Table \ref{tab:val_results} summarizes the quantitative performance on the ReXGroundingCT public validation set. Our proposed method, DAGG, consistently ranks first, maintaining a substantial performance gap over all compared baseline methods across every metric. By achieving the highest scores in both overall segmentation quality (Dice, F1) and individual grounding accuracy (Recall, HIT$_{10\%}$), DAGG firmly demonstrates the robustness of its decoupled architecture.

This superiority is rigorously validated on the unseen private test set of the official leaderboard. As presented in Table \ref{tab:test_results}, DAGG achieves state-of-the-art results in overall segmentation quality, securing the highest Global Dice, Instance Precision, and Instance F1 scores. While SAT-FT (fine-tuned) yields higher Recall and Global HIT$_{10\%}$, its drastically low precision indicates a severe tendency for over-prediction and false positives. In contrast, DAGG effectively suppresses these hallucinations, maintaining a strict balance between precision and recall to deliver highly reliable grounding masks suitable for clinical applications.

\subsection{Ablation Study}
We conduct a progressive ablation study on the ReXGroundingCT validation set to validate each core component, as summarized in Table \ref{tab:ablation}. A naive baseline using only Stage 1 localization, which indiscriminately assigns all candidate lesions to every textual finding, yields the lowest performance. Introducing the Stage 2 mapping module significantly improves all metrics by establishing semantic correspondences between text queries and visual sub-volumes. Incorporating Explicit Anatomical Guidance (EAG) further boosts grounding accuracy by providing structural priors to resolve spatial ambiguities among adjacent regions. Finally, integrating the Auxiliary Mask Refinement (AMR) module achieves our best overall results, demonstrating its effectiveness in refining coarse boundaries to ensure fine-grained spatial coherence.

\begin{table*}[t]
\centering
\caption{Progressive ablation study on the ReXGroundingCT validation set.}
\label{tab:ablation}
\begin{tabular}{ccc *{5}{>{\centering\arraybackslash}p{1.6cm}}}
\toprule
Stage 2 & EAG & AMR & Dice $\uparrow$ & F1 $\uparrow$ & Precision $\uparrow$ & Recall $\uparrow$ & HIT$_{10\%}$ $\uparrow$ \\
\midrule
\ding{55}  & \ding{55}  & \ding{55}  & 0.120 & 0.132 & 0.080 & \textbf{0.378} & 0.304 \\
\checkmark & \ding{55}  & \ding{55}  & 0.183 & 0.179 & 0.128 & 0.301 & 0.496 \\
\checkmark & \checkmark & \ding{55}  & 0.217 & 0.217 & 0.172 & 0.296 & 0.548 \\
\midrule
\checkmark & \checkmark & \checkmark & \textbf{0.250} & \textbf{0.237} & \textbf{0.196} & 0.298 & \textbf{0.565}  \\
\bottomrule
\end{tabular}
\end{table*}

\begin{figure}[t]
\includegraphics[]{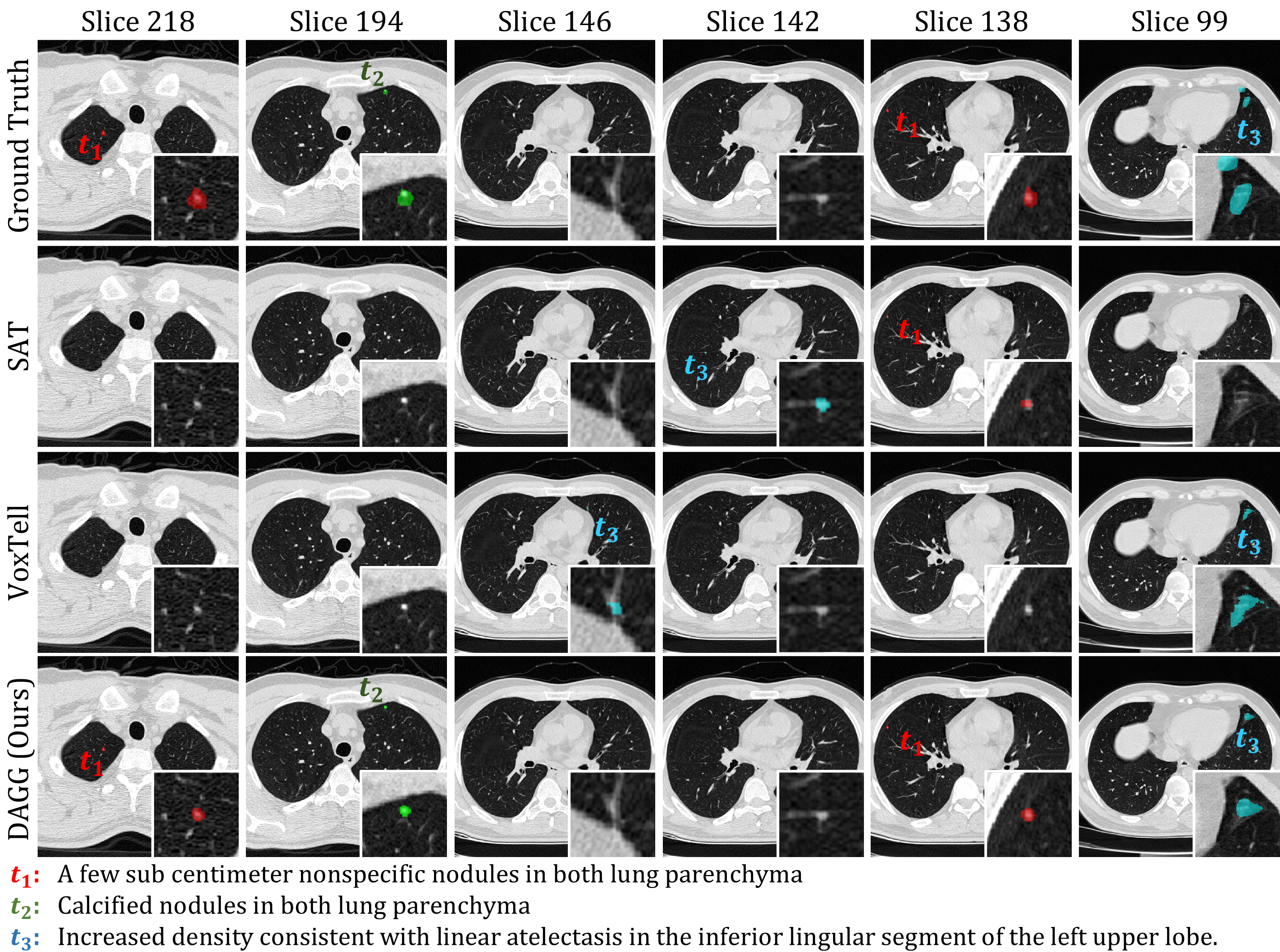}
\caption{Qualitative results for the grounding of free-text findings across multiple slices.} \label{fig2}
\end{figure}

\subsection{Qualitative Results}
Figure~\ref{fig2} visualizes the segmentation results guided by the free-text radiological findings. Compared to the baseline models, our proposed method consistently generates the most accurate segmentation masks. As depicted, our model successfully localizes the queried lesions and provides precise boundary delineation that most closely aligns with the ground truth, effectively handling complex and ambiguous anatomical structures.

\section{Conclusion}
In this paper, we proposed a novel decoupled two-stage framework for voxel-level 3D chest CT grounding that explicitly disentangles lesion localization from semantic alignment. To address the inherent complexities of free-text medical findings, we introduced an explicit anatomical guidance mechanism coupled with an independent pair-wise alignment strategy, effectively resolving spatial ambiguities within local sub-volumes. 
Extensive experiments on the ReXGroundingCT benchmark validate the effectiveness of our approach. By achieving a superior balance between precision and recall, our method achieves state-of-the-art performance on the official leaderboard. This establishes a robust and highly precise foundation for cross-modal reasoning in medical imaging.
%
%
%
\bibliographystyle{splncs04}
\bibliography{ref}
%




\end{document}